\documentclass{article}
\usepackage{etoolbox}

\PassOptionsToPackage{numbers, compress}{natbib}
 \usepackage[final, nonatbib]{nips_2018}

\usepackage{color}
\usepackage{dirtytalk}
\usepackage{bbm}
\usepackage{amsmath, dsfont,mathtools}
\usepackage{amssymb}
\usepackage{breqn}
\usepackage{subcaption}
\usepackage{url}
\usepackage{graphicx}

\usepackage{tikz}

\newcommand{\medknnMAE}{0.343}

\newcommand{\optknnMAE}{0.509}

\newcommand{\meanAUC}{0.768 }
\newcommand{\medknnAUC}{0.848}

\usepackage{graphicx,wrapfig}

\newcommand{\M}{\mathbf}
\newcommand{\Cal}{\mathcal}
\newcommand{\1}{\mathds{1}}

\newcommand{\R}{\mathbb{R}}

\usepackage[utf8]{inputenc} % allow utf-8 input
\usepackage[T1]{fontenc}    % use 8-bit T1 fonts
\usepackage{hyperref}       % hyperlinks
\usepackage{url}            % simple URL typesetting
\usepackage{booktabs}       % professional-quality tables
\usepackage{amsfonts}       % blackboard math symbols
\usepackage{nicefrac}       % compact symbols for 1/2, etc.
\usepackage{microtype}      % microtypography
\usepackage[sorting=none]{biblatex}
\newcommand{\cmmnt}[1]{}

\title{Imputation of Clinical Covariates in Time Series}

\author{
  Dimitris Bertsimas \\
  Sloan School of Management and Operations Research Center\\
  Massachusetts Institute of Technology\\
  Cambridge, MA 02142 \\
  \texttt{dbertsim@mit.edu} \\
  \And
  Agni Orfanoudaki \\
  Sloan School of Management and Operations Research Center\\
  Massachusetts Institute of Technology\\
  Cambridge, MA 02142 \\
  \texttt{agniorf@mit.edu} \\
  \AND
  Colin Pawlowski \\ 
  Sloan School of Management and Operations Research Center\\
  Massachusetts Institute of Technology\\
  Cambridge, MA 02142 \\
  \texttt{cpawlows@mit.edu} 
}

\begin{document}

\maketitle

\begin{abstract}
Missing data is a common problem in real-world settings and particularly relevant in healthcare applications where researchers use Electronic Health Records (EHR) and results of observational studies to apply analytics methods. This issue becomes even more prominent for longitudinal data sets, where multiple instances of the same individual correspond to different observations in time. Standard imputation methods do not take into account patient specific information incorporated in multivariate panel data. We introduce the novel imputation algorithm \texttt{med.impute} that addresses this problem, extending the flexible framework of \texttt{opt.impute} suggested by Bertsimas et al. \cite{optImpute}. Our algorithm provides imputations for data sets with missing continuous and categorical features, and we present the formulation and implement scalable first-order methods for a $K$-NN model.  We test the performance of our algorithm on longitudinal data from the Framingham Heart Study when data are missing completely at random (MCAR). We demonstrate that \texttt{med.impute} leads to significant improvements in both imputation accuracy and downstream model AUC compared to state-of-the-art methods.
\end{abstract}

\section{Introduction}

Despite the implementation of complex Electronic Healthcare Records (EHR) Systems, missing data are ubiquituous in clinical epidemiological research \cite{EpidResearch} posing considerable challenges in the analyses and interpretation of results and potentially weakening their validity \cite{MissHandled}. Often, those data sets contain numerous visits of the same person corresponding to various patterns of missing data. This particularity challenges state-of-the-art missing data methods that do not consider the connection of multiple observations to the same individual \cite{RRNMultTime}.
   
A variety of machine learning approaches have been introduced in the literature to deal with missing data. The simplest approach is the \texttt{mean} imputation that uses the mean of the observed values to replace the missing for the same covariate \cite{meanImpute}. Another common method called \texttt{bpca} uses the singular value decomposition of the data matrix and information from a prior distribution on the model parameters to impute the missing values \cite{bpca}. Recent studies, though, show that such methods can lead to seriously misleading results, advising to consider multiple imputation \cite{ImputeThanIgnore, meanImpute}. The latter, implemented in the package \texttt{mice} \cite{MICE}, allows for uncertainty about the missing data by creating several different plausible imputed data sets and appropriately combining results obtained from each of them \cite{MultiBehavResearch}. However, multiple imputation methods are slower and require pooling results, which may not be appropriate for certain applications. Bertsimas et al. \cite{optImpute} showed that a general optimization framework with a predictive model-based cost function can explicitly handle both continuous and categorical variables and can be used to generate single as well as multiple imputations. This optimization perspective leads to new scalable algorithms for more accurate data imputation.

The algorithms above are not tailored to multivariate time series data sets even though in time series prediction missing values and their missing patterns are often correlated with the target labels  \cite{RNNTimeImpute}.  Nevertheless, preliminary work was done in demonstrating their performance in that setting \cite{MultiAmelia}. Recurrent Neural Network approaches have also been employed \cite{RNNTimeImpute, RRNMultTime} when the missing values were imputed as part of a prediction task tailored to the particular data set. 

Given multivariate time-series data, we develop a novel imputation method that utilizes traditional optimization and machine learning techniques leading to superior performance over state-of-the-art algorithms. We formulate the problem of missing data imputation with time series information as a family of optimization problems under the \texttt{med.impute} framework.  We derive fast first-order solutions to these problems for $K$-NN which can be easily extended to SVM and tree models.  We run experiments on the Framingham Heart Study, a large-scale longitudinal clinical study, focusing on downstream models which predict 10-year risk of stroke. We demonstrate that \texttt{med.impute} leads to significant improvements in imputation accuracy and downstream model accuracy.

\section{Methods}
\subsection{Framework Definition}

We consider the single imputation problem, for which our task is to fill in the missing values of data set $\M X \in \R^{n \times p}$ with $n$ observations (rows) and $p$ features (columns).  We assume that the first $p_0$ features are continuous, with missing and known indices $\Cal M_0$, $\Cal N_0$ respectively, and that the next $p_1 = p - p_0$ features are categorical, with missing and known indices $\Cal M_1$, $\Cal N_1$ respectively.  

In addition, we assume that each observation $i$ corresponds to an individual $y_i \in \{1,\ldots,M\}$. 
For data sets with multiple observations of individuals over time, we have $M < n$.  We define $t_i \in \R^+$ as the number of (days/months/years) after a reference date that observation $i$ was recorded.  It follows that $|t_i - t_j|$ is the time difference in (days/months/years) between observations $i$ and $j$.  

For each feature $d = 1,\ldots,p$, we introduce the parameters $\alpha_d, \lambda_d$.  The first parameter $\alpha_d \in [0,1]$ is the relative weight given to the time series component of the \texttt{med.impute} objective function for variable $d$.  At the extremes, $\alpha_d = 0$ corresponds to imputing feature $d$ under the original \texttt{opt.impute} objective, and $\alpha_d = 1$ corresponds to imputing feature $d$ using each individual's time series information independently.  

The second parameter $\lambda_d \in (0,1]$ is the exponential time decay parameter for variable $d$.  We introduce this parameter so that observations from the same individual at nearby points in time will be weighted most heavily in the imputation.  For each pair of observations $i, j$, we define

\[
C_{ijd} = \left\{\begin{array}{rl}
        \lambda_d^{|t_i - t_j|}, & \text{if}~y_i = y_j, \vspace{3pt}\\
        0, & \text{otherwise}. \\
        \end{array} \right.
\]

These constants will be coefficients in the time series component of the objective function.
We learn $\alpha_d$ and $\lambda_d$ via cross-validation.  

\subsection{A $K$-NN Formulation of the Problem}
    
Given the general optimization-based imputation model suggested in \cite{optImpute}, we present an adjusted formulation that accounts for multiple instances of the same observation in time under the $K$-NN framework.

In order to weight instances of the same person in the imputation model, we will add a penalty term to the objective, with different weights $\alpha_d$ for each dimension $d = 1,\ldots,p$.  The key decision variables are the imputed continuous values $\{w_{id} \in \Cal M_0\}$ and the imputed categorical values $\{v_{id} \in \Cal M_1\}$. \\
First, define the distance between observations $i$ and $j$ as
\begin{equation}\label{eq:dist_metric}
d_{ij} := \sum_{d=1}^{p_0} (w_{id} - w_{jd})^2 + \sum_{d=p_0+1}^{p_0+p_1} \1_{\{v_{id} \neq v_{jd}\}}.
\end{equation}
Next, introduce the binary variables:
\begin{equation}\label{def:z_ij}
z_{ij} = \left\{\begin{array}{rl}
1, & \text{if $j$ is among the $K$-nearest neighbors of $i$ with respect to distance metric}~\eqref{eq:dist_metric}, \vspace{3pt}\\
0, & \text{otherwise.} \\
\end{array} \right.
\end{equation}
The \texttt{med.impute} formulation with the $K$-NN objective function is
\begin{equation}\label{eq:knn_opt}
\begin{aligned}
&\begin{aligned}
\min~~ \sum_{i\in \Cal I}\sum_{j=1}^n z_{ij}\left(\sum_{d=1}^{p_0} (1 - \alpha_d) (w_{id}-w_{jd})^2 + \sum_{d=p_0+1}^{p_0+p_1} (1 - \alpha_d) \1_{\{v_{id}\neq v_{jd}\}}\right)\\
+ \sum_{i\in \Cal I}\sum_{j=1}^n\left(\sum_{d=1}^{p_0} \alpha_d C_{ijd}(w_{id}-w_{jd})^2 + \sum_{d=p_0+1}^{p_0+p_1} \alpha_d C_{ijd}\1_{\{v_{id}\neq v_{jd}\}}\right)\\
\end{aligned}\\
&\begin{aligned}
\textrm{s.t.}~~~~~ w_{id} &= x_{id}				& (i,d) \in \Cal N_0,\\
				   v_{id} &= x_{id}				& (i,d) \in \Cal N_1,\\
				   z_{ii} &= 0		& i \in \Cal I, \\
				   \sum_{j=1}^n z_{ij}&=K		& i \in \Cal I,\\
				\M Z &\in \{0,1\}^{|\Cal I| \times n}, \\
\end{aligned}
\end{aligned}
\end{equation}
where $\Cal I = \{i : \M x_i~\text{has one or more missing values}\}$. At the optimal solution, the objective function is the sum of the distances from each point to its $K$-nearest neighbors with respect to distance metric~\eqref{eq:dist_metric}, plus the sum of the distances from each point to other observations from the same individual. We use coordinate descent \cite{BertsekasCD} with random restarts to find high quality solutions for this problem, alternatively updating the binary variables and the imputed values as in \cite{optImpute}.  

\section{Real-world Experiments}

    \subsection{Experimental Setup}
    
To test the accuracy of our method, we run a series of computational experiments on 
data from the Framingham Heart Study (FHS), a large-scale longitudinal
clinical study, focusing on downstream models which predict 10-year risk of stroke. 
We consider all individuals from the Original Cohort with 10 or more observations, which includes $M = 1,107$ unique patients.  For each patient, we take the 10 most recent observations, so the data set has $n = 11,070$ observations total.  We include $p = 13$ continuous (Body Mass Index, Systolic Blood Pressure, Hematocrit, etc.) and categorical (Gender, Smoking, etc.) covariates. 

In our experiments, we generate patterns of missing data for various percentages ranging from 10\% to 50\% under the missing completely at random (MCAR) mechanism. We take the full data set to be the ground truth. We run some of the most commonly-used and state-of-the-art methods~\cite{optImpute, meanImpute, bpca, MICE} for imputation on these data sets to predict the missing values and compare against \texttt{med.impute}. We run further experiments to evaluate the impact of these imputations on the intended downstream machine learning task, which is to predict the 10-year risk of stroke given the most recent observation from each patient.  We compare the out-of-sample performance of an $\ell_1$-regularized logistic regression model fit using the imputed data sets for various levels of missing information.  We run a second set of experiments with the missing percentage fixed at 50\%, varying the number of observations per patient in the data set from 1 to 10.  We report imputation accuracy and downstream task results for these experiments as well.

\subsection{Results}
  
\begin{figure}[h]
\centering
\begin{subfigure}{.7\textwidth}
  \centering
  \includegraphics[width=1\linewidth]{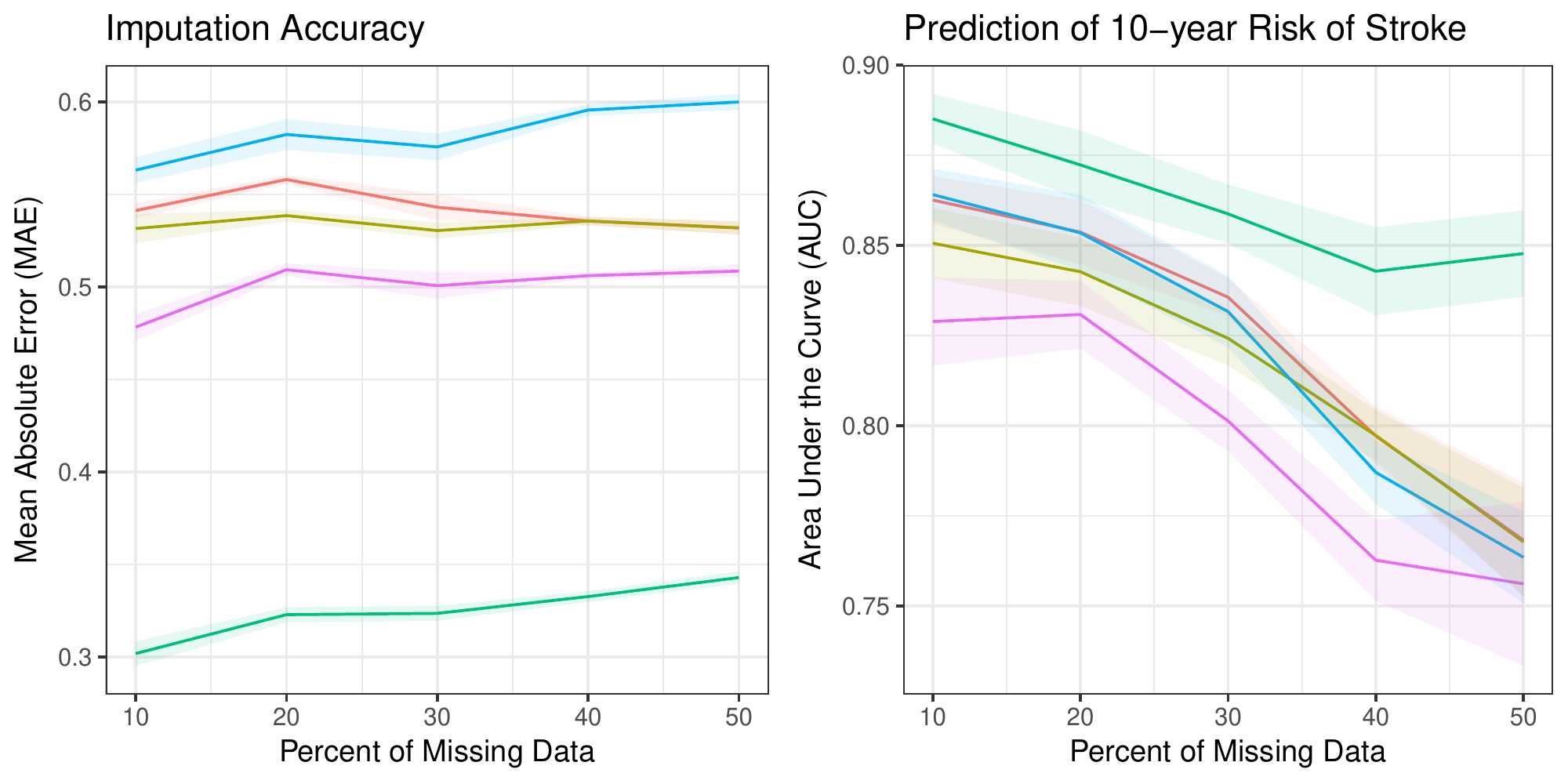}
  \label{fig:res}
\end{subfigure}%
\begin{subfigure}{.3\textwidth}
  \centering
  \includegraphics[width=0.5\linewidth]{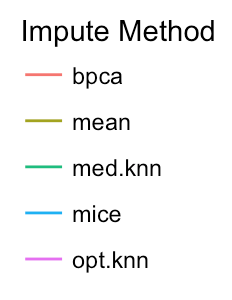}
  \label{fig:legend}
\end{subfigure}
\caption{Results from experiments on the Framingham Heart Study data set with 10 observations per patient, varying the percentage of missing data from 10\% to 50\%.}
\label{fig:FHS_missing}
\end{figure}

In Figure~\ref{fig:FHS_missing}, we show the results from the FHS experiments with varying levels of missing data.  For all missing percentages considered, \texttt{med.impute} produces the imputations with the highest
accuracy and the best performance
on the downstream classification task.  
As the percentage of missing data increases, the
relative improvement of \texttt{med.impute} over 
the reference methods increases.  At 50\% missing data, the mean absolute error (MAE) of \texttt{med.impute} is \medknnMAE, compared to \optknnMAE~for the next best method \texttt{opt.impute}.  Further, at 50\% missing data, the area under the curve (AUC) of the logistic model trained using the \texttt{med.impute} imputation is \medknnAUC, compared to \meanAUC for the next best methods \texttt{mean} and \texttt{bpca}.  These results demonstrate that \texttt{med.impute} is able to leverage time series information to gain a substantial edge over methods which ignore this information.  

\begin{figure}[h]
\centering
\begin{subfigure}{.7\textwidth}
  \centering
  \includegraphics[width=1\linewidth]{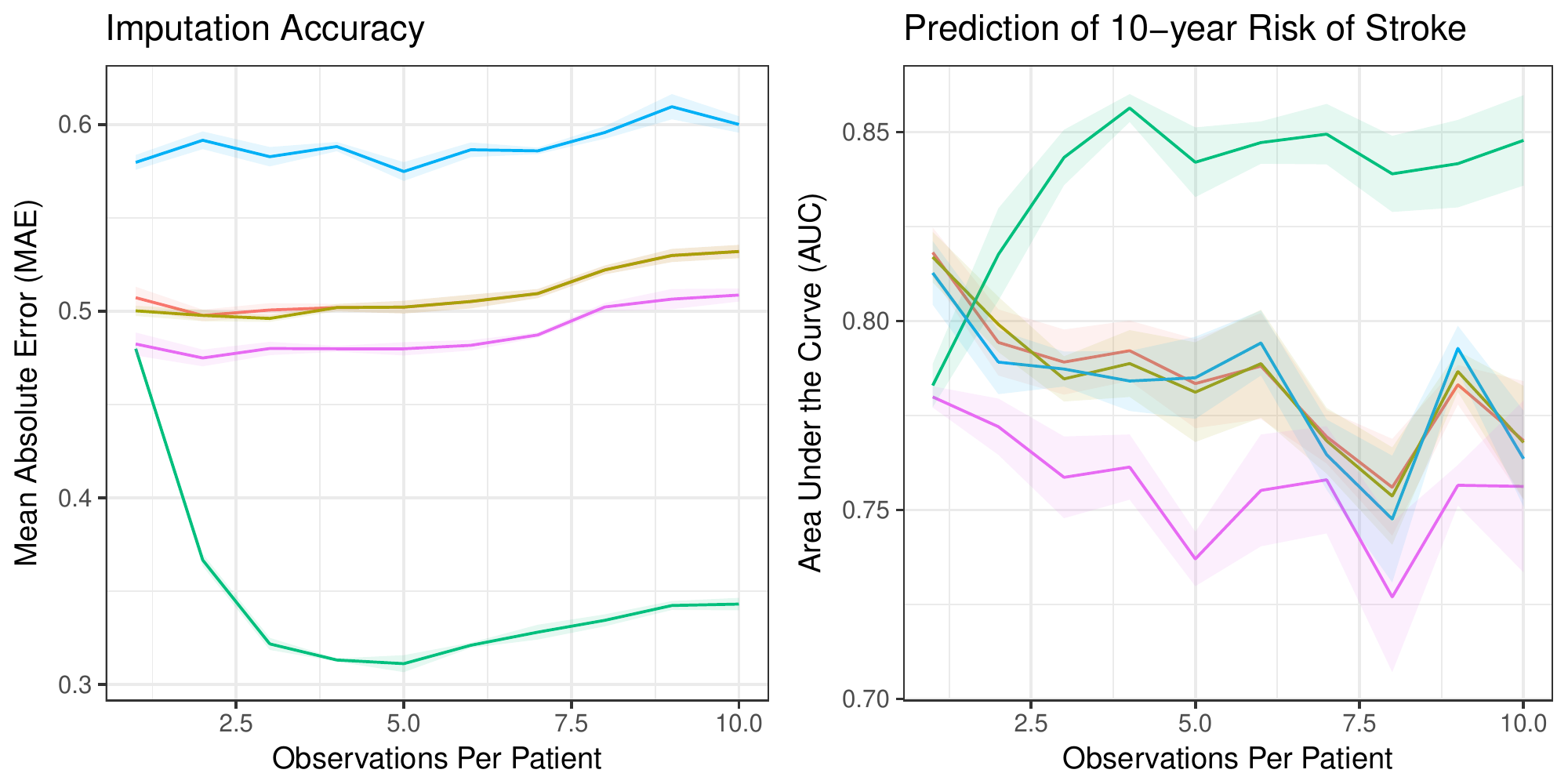}
  \label{fig:res}
\end{subfigure}%
\begin{subfigure}{.3\textwidth}
  \centering
  \includegraphics[width=0.5\linewidth]{FHD_exp1_time_10_legend.png}
  \label{fig:legend}
\end{subfigure}
\caption{Results from experiments on the Framingham Heart Study data set with 50\% missing data, varying the number of observations per patient from 1 to 10.  }
\label{fig:FHS_numobs}
\end{figure}

In Figure~\ref{fig:FHS_numobs}, we show the results from the FHS experiments with varying numbers of observations per patient (OPP).  For the experiment with OPP = $k$, we take the $k$ most recent observations for each patient.  When OPP = 1, \texttt{med.impute} is equivalent to the \texttt{opt.impute} method.  We observe that the MAE of \texttt{med.impute} decreases significantly as OPP goes up to 4-5, and then increases slightly beyond this point.  Because the observations in FHS data set occur every 2 years, this suggests that the past 6-8 years of data are most useful for imputing an individual's clinical covariates.  Similarly, the AUC from \texttt{med.impute} peaks when OPP = 4, and then levels off around 0.85.  In contrast, the AUC of the reference methods declines slightly as the OPP increases, indicating that adding more observations to the data set does not help traditional imputation methods in this case.  

\section{Conclusions}
We propose a new imputation algorithm for multivariate data in time series that yields high quality solutions using a $K$-NN framework combined with fast first-order methods. Through computational experiments with real-world data sets from the Framingham Heart Study, we show that \texttt{med.impute} yields statistically significant gains in imputation quality over state-of-the-art imputation methods, which leads to improved out-of-sample performance on downstream tasks.  In future work, we can extend this algorithm to incorporate time series information in SVM and Trees models.  

\newpage
\printbibliography
\end{document}